\documentclass[journal]{IEEEtran}
\usepackage{cite}
\usepackage{amsmath}
\usepackage{amssymb}
\usepackage{bm}
\usepackage{multicol}
\usepackage{multirow}
\usepackage{booktabs}
\usepackage{color}
\usepackage[pdftex]{graphicx}
\usepackage[setpagesize=false]{hyperref}
\usepackage{url}
\begin{document}
%
\title{Deep\! Generative\! Model\! using\! Unregularized\! Score\! for Anomaly Detection with Heterogeneous Complexity}
%
%
%

\author{Takashi Matsubara,~\IEEEmembership{Member,~IEEE,}
		Kenta Hama,
		Ryosuke Tachibana,
		and~Kuniaki~Uehara,~\IEEEmembership{Nonmember}
\thanks{T.~Matsubara, K.~Hama, R.~Tachibana and K.~Uehara are with the Graduate School of System Informatics, Kobe University, Hyogo, Japan e-mail: matsubara@phoenix.kobe-u.ac.jp.}
\thanks{Manuscript received April 19, 2005; revised August 26, 2015.}}

%
%

\markboth{Journal of \LaTeX\ Class Files,~Vol.~14, No.~8, August~2015}%
{Matsubara \MakeLowercase{\textit{et al.}}: Anomaly Detection with Heterogeneous Complexity using a Deep Generative Model with an Unregularized Score}
%



\maketitle

\begin{abstract}
Accurate and automated detection of anomalous samples in a natural image dataset can be accomplished with a probabilistic model for end-to-end modeling of images.
Such images have heterogeneous complexity, however, and a probabilistic model overlooks simply shaped objects with small anomalies.
This is because the probabilistic model assigns undesirably lower likelihoods to complexly shaped objects that are nevertheless consistent with set standards.
To overcome this difficulty, we propose an unregularized score for deep generative models (DGMs), which are generative models leveraging deep neural networks.
We found that the regularization terms of the DGMs considerably influence the anomaly score depending on the complexity of the samples.
By removing these terms, we obtain an unregularized score, which we evaluated on a toy dataset and real-world manufacturing datasets.
Empirical results demonstrate that the unregularized score is robust to the inherent complexity of samples and can be used to better detect anomalies.
\end{abstract}


%
\IEEEpeerreviewmaketitle

\section{Introduction}
Image-based anomaly detection has recently attracted considerable attention in the field of machine learning.
This technique can be used to detect pedestrians behaving abnormally from surveillance video in order to prevent accidents~\cite{Leach2014,Ribeiro2017}, or to detect lesions in medical images to provide early diagnosis~\cite{Schlegl2017}.
In manufacturing plants, moreover, image-based anomaly detection can reject products not coincident with set standards.
Accurate and automatic detection of anomalous products thus reduces workload and inspection costs, and can improve product reliability.
Since the size of accessible data is ever-increasing, manually labeling all of the samples is either impossible or prohibitively expensive.
Furthermore, machine learning techniques cannot practically rely on extensive adjustments by experts.
Consequently, a practical demand for unsupervised learning arises.

Unsupervised anomaly detection is often based on a probabilistic model $p(x)$ of the target domain (see \cite{Chandola2009} for a survey).
Since training samples are expected to contain few if any anomalies, a probabilistic model $p(x)$ trained with the training samples assigns lower likelihoods to anomalous test samples.
Because it is difficult to model natural images directly, many studies proposed extracting features from such images using unsupervised methods such as principal component analysis (PCA)~\cite{Kim2009,Mahadevan2010,Saligrama2012,Leach2014,Li2017}.
In recent years, generative models leveraging deep neural networks (DNNs) have been proposed, called \emph{deep generative models} (DGMs)~\cite{Kingma2014,Kingma2014a,Sohn2015,Maaloe2015}.
A DGM can build a generative model of natural images in an end-to-end manner, achieving a certain measure of success with unsupervised anomaly detection~\cite{Zhai2016,Suh2016,Lopez-Martin2017,Chalapathy2017,Zhou2017,Ribeiro2017}.

\begin{figure}[tb]
	\centering
	\includegraphics[width=3.5in,page=3]{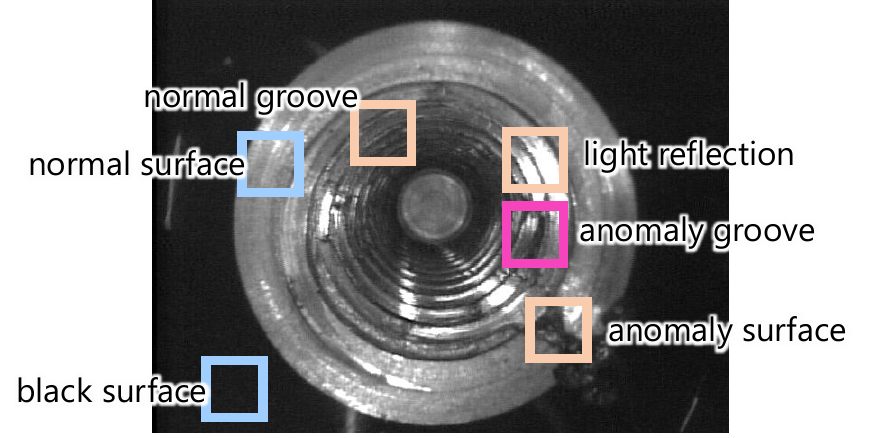}
	\caption{Conceptual image of heterogeneous complexity.
	The screw holes provide a wide variety of image patches, resulting in lower likelihoods than the anomaly along the flat surface.
	}\label{fig:punch}
	\vspace*{-5mm}
\end{figure}

However, anomaly detection of machine components with complex structures remains challenging because of the wide variety of their image patches.
Casting products have flat and smooth surfaces but also curved parts and screw holes.
As shown in Fig.~\ref{fig:punch}, a screw hole image is inherently more sensitive to the variability of its production (e.g., the starting point of the screw groove), the imaging direction, and the illumination condition.
A screw hole is more \emph{complex} than a flat surface, and hence, it produces a wide variety of image patches with lower likelihoods.
A probabilistic model $p(x)$ thus risks detecting a normal screw groove as an anomaly instead of a flat surface with a small scratch.
Medical image datasets encounter the same issue: images depict various types of tissues, some of which have complex structures and appear less frequently.
Generally, the differences among normal samples belonging to a complex class are higher than between normal and anomalous samples belonging to an ordinary class.
In this paper, we refer to this property as \emph{heterogeneous complexity}.
Not limited to these real-world datasets, typical benchmark datasets such as the MNIST handwritten digit database~\cite{MNIST} and the Street View House Numbers (SVHN) Dataset~\cite{Netzer2011} are heterogeneously complex; some digits such as ``8'' have more complex shapes and a wider variety than others such as ``1'' and ``7''.
This is a plausible reason for why DNN-based anomaly detection has not worked as expected in recent works~\cite{Aytekin2018,Chen2018c}.

To overcome this difficulty, we propose an \emph{unregularized score} as a novel anomaly score for the DGM~\cite{Kingma2014}.
As its name implies, the unregularized score is the anomaly score of the DGM without regularization terms.
We demonstrate that even though the regularization terms of the DGM contribute to building a better model during the training phase, they considerably influence the anomaly score depending on the complexity of samples and they are not suited for anomaly detection.
By removing these terms during the detection phase, the DGM is rendered robust to the inherent complexity of samples and is less likely to reject samples that have complex shapes but are coincident with set standards.
We evaluated DGMs with the unregularized score on a toy dataset based on the SVHN Dataset~\cite{Netzer2011} and two real-world datasets of machine component images.
The results confirm that a DGM with the proposed unregularized score outperforms its counterparts and conventional approaches (viz., the Gaussian mixture model~\cite{Saligrama2012,Lu2014,Li2017} and Isolation Forest~\cite{Liu2012}) by large margins.

The remainder of this paper is organized as follows.
We introduce previous works on anomaly detection and DGMs in Section~\ref{sec:related}.
In Section~\ref{sec:methods}, we propose the unregularized score for the DGMs, and conceptually explain the heterogeneously complex dataset and the reason why the proposed unregularized score works well for such datasets.
Section~\ref{sec:data} describes the toy dataset and two real-world datasets of machine component images used in our experiments.
Section~\ref{sec:results} introduces the detailed experimental setting of the DGM with our proposed unregularized score and competitive methods as well as their numerical results.
In Section~\ref{sec:discussion}, we verify our concept of the heterogeneous complexity and the efficiency of the proposed unregularized score.

Limited preliminary results can be found in a conference proceeding~\cite{Matsubara2018WCCI}.

\begin{figure*}[tb]
	\centering
	\includegraphics[width=4.8in,page=2]{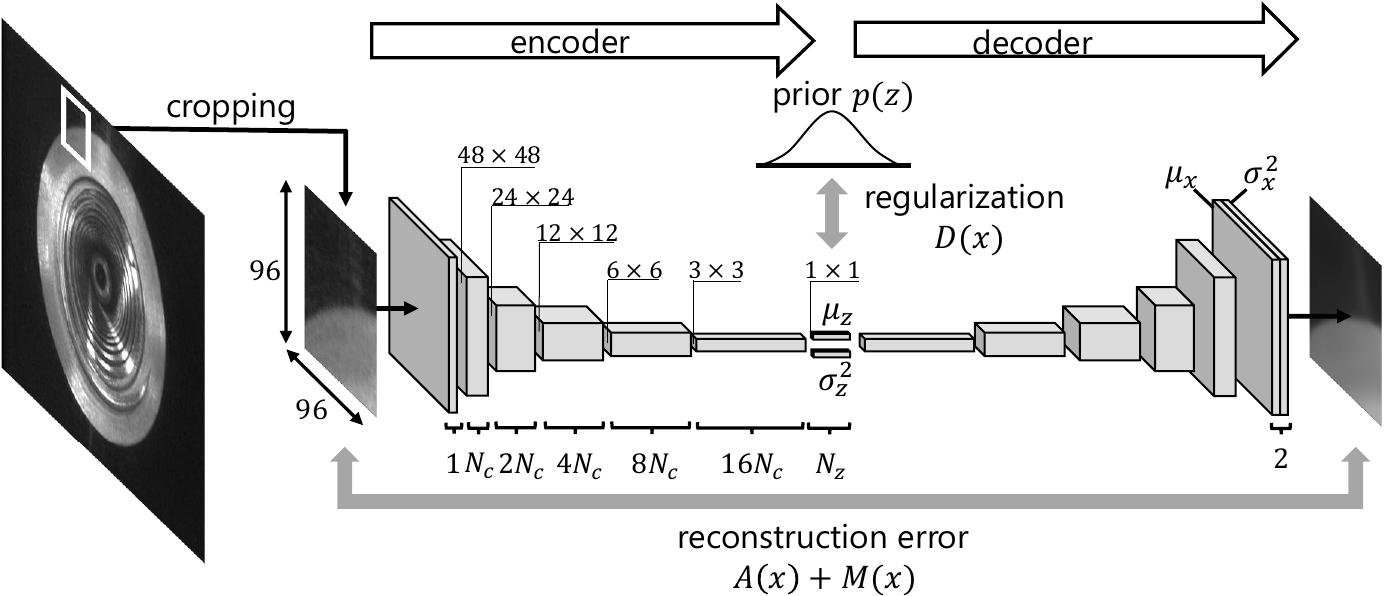}
	\caption{Diagram of the variational autoencoder (VAE) implemented on convolutional neural networks (CNNs).}\label{fig:cnn}
\end{figure*}

\section{Related Works}\label{sec:related}

\subsection{Unsupervised Anomaly Detection}
Unsupervised anomaly detection often employs a probabilistic model (see \cite{Chandola2009} for a survey).
A probabilistic model is trained with training samples and then applied to test samples.
Since the training samples are assumed to contain few or no anomalies, an anomaly test sample has a low probability of being generated from the model.

A Gaussian mixture model (GMM) is one type of generative model used for anomaly detection~\cite{Saligrama2012,Lu2014,Li2017}.
A GMM assumes that each sample $x$ is in the data space $\mathcal X\subset \mathbb R^{N_x}$ and belongs to one of the hidden classes $z\in\{1,2,\dots,N_z\}$.
Each hidden class $z$ has a multivariate Gaussian distribution as its base distribution.
Then, the GMM is expressed as
\begin{equation}\setlength\arraycolsep{1pt}
	\begin{array}{rcl}
		p_\theta(x)
		&=&\!\displaystyle\sum_{k=1}^{N_z} w_k p_\theta(x|z=k)\\
		&=&\!\displaystyle\sum_{k=1}^{N_z} w_k \frac{1}{\sqrt{(2\pi)^{N_x}\!|\!\Sigma_k\!|}}\exp\!\left(\!-\frac{1}{2}(x\!\!-\!\!\mu_k)^T\Sigma_k^{-1}(x\!\!-\!\!\mu_k)\!\right)\!\!,\\
	\end{array}
\end{equation}
where $\theta$ is a set of parameters, $w_k$ denotes the mixture weight of the hidden class $k$ satisfying $\sum_{k=1}^{N_z} w_k=1$, and $\mu_k$ and $\Sigma_k$ respectively denote the mean vector and the covariance matrix of the multivariate Gaussian distribution of the hidden class $k$.
GMMs are often trained using the Expectation-Maximization (EM) algorithm to minimize the negative log-likelihood (NLL):
\begin{equation}\setlength\arraycolsep{1pt}
	\begin{array}{rcl}
		\mathcal L(x)
		&=&\displaystyle-\log p_\theta(x)
	\end{array}
\end{equation}
This is the typical anomaly score for a GMM.

Because it is difficult to model natural images directly, many studies employ unsupervised dimension-reduction methods such as principal component analysis (PCA), saliency detection, and optical flow to extract features from such images~\cite{Kim2009,Mahadevan2010,Saligrama2012,Leach2014,Lu2014,Li2017}.
Then, probabilistic models are employed to model the extracted features and detect anomalies.

\subsection{Probabilistic Model on Neural Networks}
To leverage the flexibility of deep neural networks (DNNs), probabilistic models implemented on DNNs have been investigated, e.g., the Boltzmann machine, neural variational inference learning, and generative stochastic network~\cite{Hinton2007,Mnih2014a,Bengio2013c}.
Recent studies have proposed a \emph{deep generative model} (DGM) as an alternative implementation~\cite{Kingma2014}.
A DGM can build a model of natural images without feature extractions, i.e., in an end-to-end manner.
This is considered to be a more general and robust model for training than other DNN-based generative models because a DGM can implement a wide variety of probabilistic models and is less dependent on Monte Carlo sampling~\cite{Kingma2014a,Sohn2015,Maaloe2015}.

We introduce the simplest version of a DGM, viz., the variational autoencoder (VAE)~\cite{Kingma2014}.
We consider a probabilistic model $p_\theta(x)$ of a sample $x$ in the data space $\mathcal X\subset \mathbb R^{N_x}$ with a latent variable $z$ in a latent space $\mathcal Z\subset \mathbb R^{N_z}$ for $N_z<N_x$:
\begin{equation}
	p_\theta(x)=\int_z p_\theta(x|z)p(z),
\end{equation}
where $\theta$ is a set of parameters, and $p(z)$ is the prior distribution of the latent variable $z$.
Based on the variational method~\cite{Murphy2012}, the model evidence $\log p_\theta(x)$ is bounded using an inference model $q_\phi(z|x)$ parameterized by $\phi$ as
\begin{equation}\setlength\arraycolsep{1pt}
	\begin{array}{rcl}
	\log p_\theta(x)
	&=&\displaystyle \mathbb E_{q_\phi(z|x)}\left[\log\frac{p_\theta(x,z)}{p_\theta(z|x)}\right]\\[4mm]
	&=&\displaystyle \mathbb E_{q_\phi(z|x)}\left[\log\frac{p_\theta(x,z)}{q_\phi(z|x)}\right]+D_{KL}(q_\phi(z|x)||p_\theta(z|x))\\[4mm]
	&\ge&\displaystyle \mathbb E_{q_\phi(z|x)}\left[\log\frac{p_\theta(x,z)}{q_\phi(z|x)}\right]\\[4mm]
	&=&-D_{KL}(q_\phi(z|x)||p(z)) + \mathbb E_{q_\phi(z|x)}\left[\log p_\theta(x|z)\right]\\[1mm]
	&=:&- \mathcal L(x),
	\end{array}\label{eq:scanlikelihood}
\end{equation}
where $D_{KL}(\cdot||\cdot)$ is the Kullback--Leibler divergence and $- \mathcal L(x)$ is the evidence lower bound (ELBO).
The negative ELBO $\mathcal L(x)$ is the objective function to be minimized.

The VAE implements these probabilistic models $p_\theta(x|z)$ and $q_\phi(z|x)$ on DNNs respectively called the \emph{encoder} and \emph{decoder} using the \emph{reparameterization trick}.
The encoder accepts a data sample $x$ and infers the parameters of the variational posterior $q_\phi(z|x)$ instead of a point estimate of the latent variable $z$.
Then, the Kullback--Leibler divergence $D_{KL}(q_\phi(z|x)||p(z))$ can be calculated.
The decoder accepts a latent variable $z$ and then outputs the parameters of the conditional probability $p_\theta(x|z)$ of the data sample $x$.
This also enables the calculation of the conditional log-likelihood $\log p_\theta(x|z)$.
The expectation $\mathbb E_{q_\phi(z|x)}\left[\log p_\theta(x|z)\right]$ is calculated by Monte Carlo sampling from the variational posterior $q_\phi(z|x)$.
In the original implementation~\cite{Kingma2014}, the latent variable $z$ is sampled once each iteration during training.
In the test phase, the maximum a posteriori (MAP) estimate of the latent variable $z$ (i.e., the mean vector $\mu_{z}$) is used in place of Monte Carlo sampling for simplicity.
Previous studies confirmed that this simplification does not seriously degrade the performance~\cite{Kingma2014,Burda2015a,Maaloe2015}.
For anomaly detection, the VAEs use the ELBO $-\mathcal L(x)$ in place of the log-likelihood $\log p_\theta(x)$.
That is, the negative ELBO $\mathcal L(x)$ is the anomaly score for the VAE~\cite{Suh2016,Lopez-Martin2017}.

\section{Methods}\label{sec:methods}

\subsection{Unregularized Score for Variational Autoencoder}
In this section, we propose the \emph{unregularized score} for anomaly detection by the VAE (see also Fig.~\ref{fig:cnn}).

Following the original study~\cite{Kingma2014}, the prior distribution $p(z)$ is set to a standard multivariate Gaussian distribution.
Each variational posterior $q_\phi(z|x)$ and the conditional probability $p_\theta(x|z)$ is modeled as a multivariate Gaussian distribution with a diagonal covariance matrix.
The output of the encoder is a pair: a mean vector $\mu_{z}$ and a standard deviation vector $\sigma_{z}$ of the variational posterior $q_\phi(z|x)=\mathcal N(\mu_z(x),\mathrm{diag}(\sigma_z(x)))$.
The output of the decoder is also a pair: a mean vector $\mu_{x}$ and a standard deviation vector $\sigma_{x}$ of the conditional probability $p_\theta(x|z)=\mathcal N(\mu_x(z),\mathrm{diag}(\sigma_x(z)))$.
As described above, the MAP estimate $\mu_z$ of the latent variable $z$ is used for anomaly detection instead of Monte Carlo sampling from the variational posterior $q_\phi(z|x)$.
Then, the negative ELBO $\mathcal L(x)$ can be rewritten as
\begin{equation}\setlength\arraycolsep{2pt}
	\begin{array}{rcl}
	\mathcal L(x)
	&=&D_{KL}(q_\phi(z|x)||p(z))-\log p_\theta(x|\mu_z)\\[1mm]
	&=&D(x)+A(x)+M(x)
	\end{array}
\end{equation}
where
\begin{equation}\setlength\arraycolsep{2pt}
	\begin{array}{rcl}
	D(x)&=&\displaystyle\sum_{j=1}^{N_z} \frac{1}{2}(-\log \sigma_{z_j}^2-1+\sigma_{z_j}^2+\mu_{z_j}^2),\\[4mm]
	A(x)&=&\displaystyle \sum_{i=1}^{N_x}\frac{1}{2}\log2\pi\sigma_{x_i}^2\bigg|_{z=\mu_z},\\[4mm]
	M(x)&=&\displaystyle \sum_{i=1}^{N_x}\frac{1}{2}\frac{(\mu_{x_i}-x_i)^2}{\sigma_{x_i}^2}\bigg|_{z=\mu_z},\\[2mm]
	p(z)&=&\mathcal N(\mathbf 0,I).
	\end{array}
\end{equation}
With regard to neural networks, $A(x)+M(x)$ is referred to as a reconstruction error, and $D(x)$ is a regularization term.
$A(x)$ corresponds to the logarithm of the normalizing constant, which makes the integral of the probability density function of the Gaussian distribution $p_\theta(x|z)$ equal to 1.
$M(x)$ is apparently similar to the square of Mahalanobis' distance or the normalized Euclidean distance.

Here, we propose an alternative score $M(x)$ in place of the negative ELBO $\mathcal L(x)=D(x)+A(x)+M(x)$.
Because the log-normalizing constant $A(x)$ and the regularization term $D(x)$ are removed, we call this score $M(x)$ an \emph{unregularized score}.
Note that the objective function of the VAE during the training phase is still the negative ELBO $\mathcal L(x)=D(x)+A(x)+M(x)$.
Below, we examine alternative scores: viz., $D(x)$ and $A(x)$.

\subsection{Unregularized Score for a Gaussian Mixture Model}
For comparison, we also introduce the unregularized score for a GMM.
When the hidden class $z$ that a sample $x$ belongs to is determined by the MAP estimate, as with the VAE, the NLL of the sample $x$ is
\begin{equation}\setlength\arraycolsep{1pt}
	\begin{array}{rcl}
		- \log p(x|z=k)=D(x)+A(x)+M(x)
	\end{array}
\end{equation}
where
\begin{equation}\setlength\arraycolsep{2pt}
	\begin{array}{rcl}
	D(x)&=&\displaystyle-\log w_k,\\
	A(x)&=&\displaystyle\frac{1}{2}\log{({2\pi})^{N_x}|\Sigma_k|},\\[2mm]
	M(x)&=&\displaystyle\frac{1}{2}(x-\mu_k)^T\Sigma_k^{-1}(x-\mu_k),
	\end{array}
\end{equation}
and $k=\arg\max_z p(z|x)$.
Then, the unregularized score of the GMM is $M(x)$.
When the number of the hidden classes $N_z$ of the GMM is set to one, the unregularized score $M(x)$ is equivalent to half the square Mahalanobis' distance $(x-\mu_k)^T\Sigma_k^{-1}(x-\mu_k)$~\cite{Chandola2009}.

\subsection{Concept of the Unregularized Score}\label{sec:concept}
We assume datasets composed of a wide variety of clusters (subpart groups in manufacturing machine components) where several clusters are far more complex than others, as shown in Fig.~\ref{fig:punch}.
For example, machine components have flat surfaces, curved parts, and screw holes.
The MNIST and SVHN datasets~\cite{MNIST,Netzer2011} are composed of complexly shaped digits like ``8'' and simply shaped digits like ``1'' and ``7''.
We call this property the heterogeneous complexity.

If a probabilistic model has a limited expression ability, it has difficulty modeling complexly shaped samples and detects them as anomalies mistakenly.
Complexly shaped samples have a wider variety than simply shaped samples and they are distributed over a larger range in the data space $\mathcal X$.
Hence, the values at the probabilistic density function (PDF) at complexly shaped samples are lower.
That means that even an ideal probabilistic model assigns lower likelihoods to complexly shaped samples and detects them as anomalies.

The GMM detects an outlier (i.e., a sample with a large reconstruction error $M(x)+A(x)$) as an anomaly, or the GMM classifies an anomalous sample into an anomaly class $z$, where the anomaly class has a low mixture weight $w_z$ and the negative log-mixture weight $D(x)$ is thus large.
The GMM estimates the NLL $\mathcal L(x)$ of a sample directly using the generative model $p(x)=\sum_z p_\theta(x|z)p(z)$, and hence a minor anomaly in a given sample $x$ affects the NLL $\mathcal L(x)$.

On the other hand, the VAE infers the latent variable $z$ to calculate the negative ELBO $\mathcal L(x)$.
The VAE is usually \emph{undercomplete}: the latent space $\mathcal Z$ has fewer dimensions than the data space $\mathcal X$ (i.e., $N_z<N_x$).
This implies that, in order to compress the samples, the encoder of the VAE learns the most salient features in the given samples and ignores features that appear less frequently.
When an anomalous sample $x_a$ includes a minor anomaly but is very similar to another normal sample $x$ obtained from the same cluster, the encoder overlooks this minor anomaly and projects the anomalous sample $x_a$ to a location near the normal sample $x$.
Hence, the regularization term $D(x)$ does not distinguish the anomalous sample $x_a$ from the normal sample $x$.

Likewise, the log-normalizing constant $A(x)$ does not distinguish the anomalous sample $x_a$ from the normal sample $x$ since it also depends on the latent variable $z$, like the regularization term $D(x)$.
Even worse, the log-normalizing constant $A(x)$ is sensitive to the heterogeneous complexity of samples.
Provided that a sample $x$ is simply shaped and the VAE reconstructs the sample $x$ accurately (i.e., outputs the estimated mean vector $\mu_x$ close to the sample $x$), an appropriate standard deviation $\sigma_{x}$ is near-zero, which minimizes the log-normalizing constant $A(x)$.
However, if the sample $x$ is complexly shaped and the VAE cannot estimate the appropriate mean vector $\mu_x$, it outputs a large standard deviation $\sigma_{x}$ to prevent the square normalized distance $M(x)$ from exploding.
The VAE is trained to balance the two criteria $A(x)$ and $M(x)$ by adjusting the standard deviation $\sigma_{x}$ depending on the uncertainty of the reconstruction.
In other words, the standard deviation $\sigma_{x}$ (and hence the log-normalizing constant $A(x)$) represents the complexity of the sample $x$ rather than the abnormity.

\begin{figure}[t]
	\centering
	\includegraphics[width=1.5in]{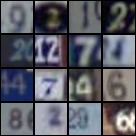}\hspace*{8mm}
	\includegraphics[width=1.5in]{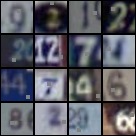}
	\vspace*{-2mm}
	\caption{Test samples of the toy dataset; (left panel) normal samples and (right panel) anomalous samples.
	}
	\label{fig:svhn}
	\vspace*{-3mm}
\end{figure}

The VAE is a heteroscedastic model like the ARCH model~\cite{Engle1982} since the standard deviation $\sigma_x$ depends on the sample $x$.
A sample $x$ can be rewritten as $x=\mu_x+S\epsilon$, where $SS^T=diag(\sigma_x^2)$ and $\epsilon\sim\mathcal N(\mathbf{0},\mathbf{I})$.
Then, the square normalized distance is rewritten as $M(x)=\frac{1}{2}||\epsilon||^2_2$, which is the negative log-likelihood of the noise term $\epsilon$ plus a constant term.
This functions as an anomaly score of the given sample $x$ independent of heteroscedasticity (i.e., heterogeneous complexity).

In the following sections, we describe our evaluation of the VAE with the unregularized score $M(x)$ and this concept.

\section{Data Acquisition}\label{sec:data}
\subsection{Toy Dataset}
First, we evaluated the unregularized score on a toy dataset made from the Street View House Numbers (SVHN) Dataset~\cite{Netzer2011}.
The dataset is composed of $32 \times 32$ RGB images depicting one of the digits obtained from house numbers in Google Street View images.
We used 73,257 training samples and did not use the extra subset to train GMMs using our computational resources.
We randomly selected 1 \% of samples as anomalous samples by randomly erasing their $4\times 4$ subregions.
We duplicated 26,032 test samples, keeping half as normal samples and randomly erasing subregions of the other half to obtain anomalous samples (see Fig.~\ref{fig:svhn}).

The toy dataset was a mixture of the ten types of digits: some digits, such as ``8'', have more complex shapes and a wider variety than others, such as ``1'' and ``7''.
Hence, this dataset was appropriate to evaluate the robustness of the proposed score to the heterogeneous complexity.

\subsection{Real-World Manufacturing Datasets}
\begin{figure}[t]
	\centering
	\includegraphics[width=1.72in]{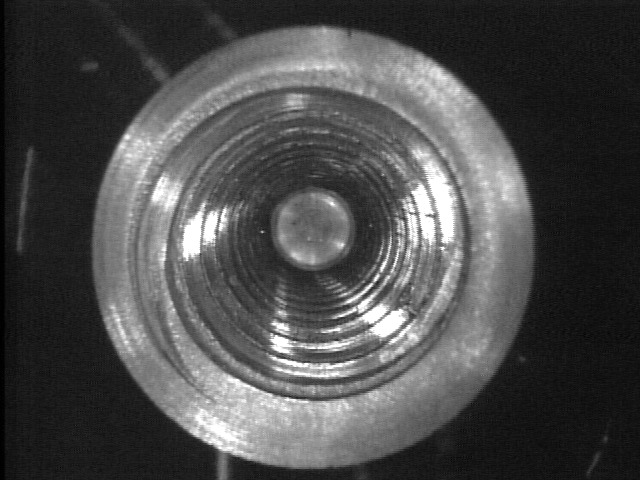}\\[2mm]
	\includegraphics[width=1.72in]{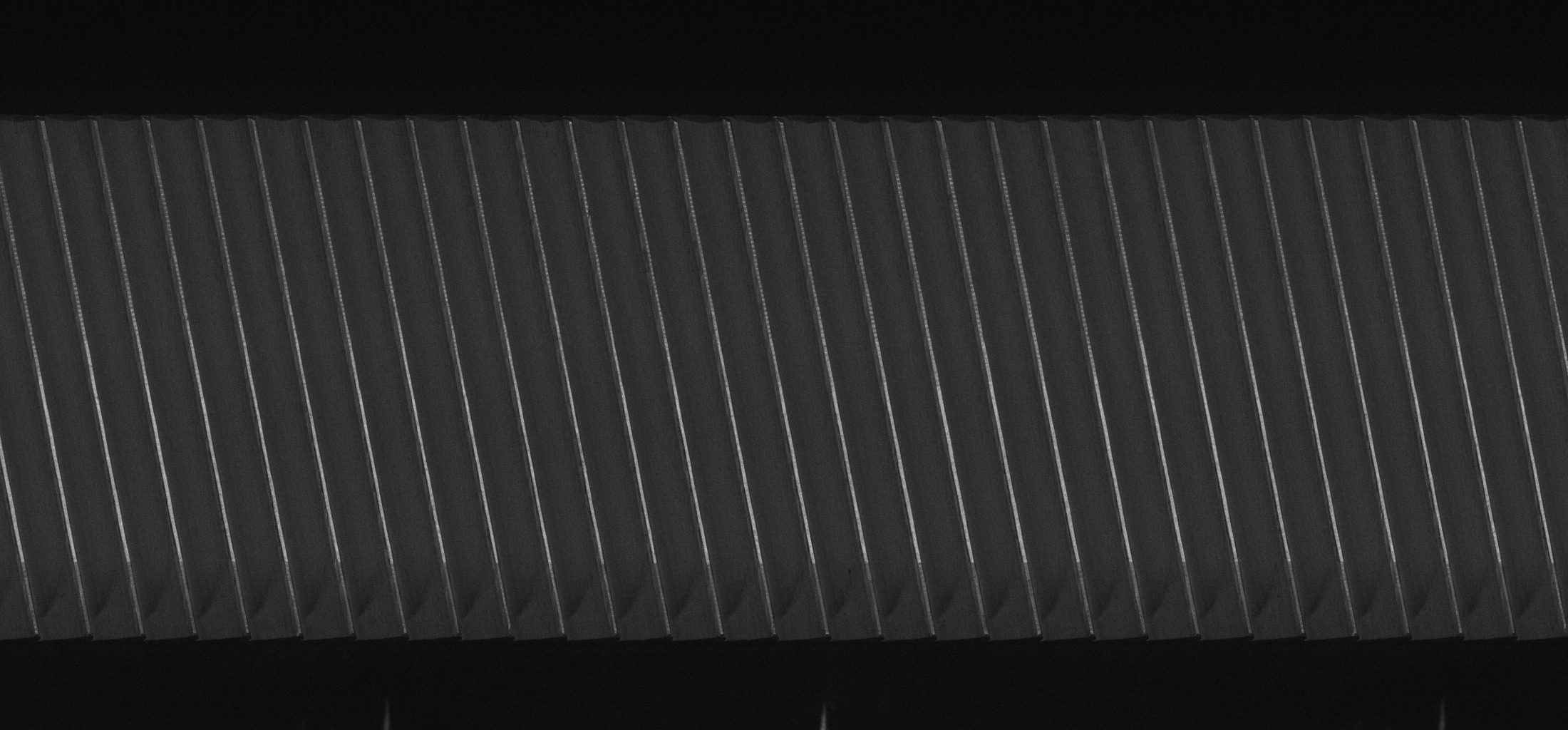}
	\caption{Samples from manufacturing datasets.
	(top panel) Sample from the screw dataset.
	(bottom panel) Sample from the gear dataset.
	}
	\label{fig:exaples}
\end{figure}

We evaluated the unregularized score on the image datasets of machine components obtained from production lines owned by AISIN AW CO., LTD.
The \emph{screw dataset} is a dataset of $640 \times 480$ grayscale images of screw holes (see the top panel of Fig.~\ref{fig:exaples}).
Each image shows a screw hole at the center and the flat black surface surrounding the screw hole.
It also shows the camera enclosure at the right and left ends.
Typical anomalies in the screw dataset are discontinuities such as cracks, seams, and porosity, caused by entrained gas and shrinkage of the material.
Such discontinuities deteriorate the casting strength, and flakes can cause problems.
The image patches of this dataset form two or more clusters: screw grooves, flat surfaces, bottoms of screw holes, and so on.
The screw dataset is composed of 12,406 training samples and 995 test samples (including 888 normal samples and 107 anomalies).
The test samples were labeled by experts from AISIN AW CO., LTD.~and used for performance evaluation, and the training samples were unlabeled and used to adjust the parameters.
The training samples were expected to contain the same rate of anomalies as the test samples.

The \emph{gear dataset} is a dataset of $2,200 \times 1,024$ grayscale images of the tooth surfaces of gear wheels (see the bottom panel of Fig.~\ref{fig:exaples}).
One of the typical anomalies in the gear dataset is an unpolished casting surface, which results from insufficient adjustment of the brushes.
This produces undesired friction with other gears.
A chipped tooth is another typical anomaly.
Since the gear teeth are aligned periodically, a sample in the gear dataset is relatively homogeneous but it is composed of the tooth, the roots, and edges.
The gear dataset is composed of 3,260 labeled samples (including 3,150 normal samples and 110 anomalies).

During training, a training sample was randomly cropped to sizes of $96\times 96$.
In the test phase, a test sample was sequentially cropped to the same size with a stride of $16\times 16$.
If the anomaly score of at least one image patch exceeded a threshold, the test sample was considered an anomaly.

\section{Experiments and Results}\label{sec:results}
\subsection{Deep Generative Models}
We implemented the VAE depicted in Fig.~\ref{fig:cnn} using PyTorch v0.4.1~\cite{Paszke2017}.
We evaluated the proposed unregularized score $M(x)$ and the ordinary anomaly score $\mathcal L(x)=D(x)+A(x)+M(x)$.

The VAE has an encoder composed of $N_{conv}$ convolution layers and one fully-connected layer.
We set the depth $N_{conv}$ to four for the toy dataset and to five for the real-world datasets.
The encoder accepts an image of $N_{size} \times N_{size}$ pixels.
The $n$-th convolution layer had a $4\times 4$ kernel and a stride of $2$ and outputted a feature map of $N_c\times 2^{n-1}$ channels.
Each convolution layer was followed by batch normalization~\cite{Ioffe2015} and the ReLU activation function~\cite{Nair2010}.
The fully-connected layer outputted $2\times N_z$ units, followed by the identity function as the activation function.
$N_z$ units were used as the mean vector $\mu_{z}$ and the other units were considered to represent the log-variance $\log \sigma_z^2$.
This made it easier to calculate the Kullback--Leibler divergence $D_{KL}(q_\phi(z|x)||p(z))$.
The decoder had a structure paired up with the encoder.
The output of the decoder was an image of a pair of channels: the mean vector $\mu_{x}$, and the log-variance $\log \sigma_x^2$.
The VAE was trained using the Adam optimizer~\cite{Kingma2014b} with parameters of $\alpha=10^{-3}$, $\beta_1=0.9$, and $\beta_2=0.999$ and a weight decay of $0.0001$.
We fixed the number $N_c$ of channels of the first feature map to $N_c=32$ for the toy dataset and selected it from $N_c\in\{16,32,64\}$ for the real-world manufacturing datasets.
We also selected the dimension number $N_z$ of the latent space $\mathcal Z$ from $\{2,5,10,20,50,100,\dots\}$.
All the other conditions followed the original study~\cite{Kingma2014} and previous studies~\cite{Zhai2016,Suh2016,Lopez-Martin2017,Chalapathy2017,Zhou2017,Ribeiro2017}.

For comparison, we also evaluated the autoencoder (AE).
The AE is a simpler version of the VAE and has also been used for anomaly detection~\cite{Zhou2017,Chalapathy2017}.
The AE has an encoder and decoder outputting point estimates of the latent variable $z$ and the reconstruction $\tilde x$, respectively.
The objective function and the anomaly score were the mean-squared error between the sample $x$ and the reconstruction $\tilde x$, i.e., $\mathcal L(x)=\frac{1}{N_x}\sum_{i=1}^{N_x} (x_i-\tilde x_i)^2$.
In other words, the AE does not need Monte Carlo sampling, has no regularization term $D(x)$, and has a constant standard deviation $\sigma_{x_i}$.
The other conditions were the same as those for the VAE.

\subsection{Models for Comparison}
We implemented the GMM and an Isolation Forest using scikit-learn v0.19.1~\cite{Pedregosa2012}.

We trained the GMM of full covariance matrices using the EM algorithm and used the negative log-likelihood $\mathcal L(x)$ and the unregularized score $M(x)$ as the anomaly score.
We selected the number of hidden classes from $N_z\in\{2,5,10,20,50,100,\dots\}$.
When applying the GMM to the real-world manufacturing datasets, we prepared ten image patches for each training sample to ensure convergence in a reasonable time.
In addition, following previous studies~\cite{Kim2009,Mahadevan2010,Saligrama2012,Leach2014,Li2017}, we extracted an $N_h$-dimensional feature from each patch using PCA.
We selected the number of principal components from $N_h\in\{20,50,100,200,500\}$.

The Isolation Forest~\cite{Liu2012} is a Random Forest trained to separate outliers.
We built each $N_e=1000$ base estimators (decision trees) using 256 image patches randomly cropped from all training samples.
We selected the estimated contamination ratio $r$ of anomalies from $r\in\{\dots,0.01,0.02,0.05,0.1,0.2,0.5\}$.

We also examined a one-class SVM~\cite{Scholkopf2001}, a support vector machine trained to separate outliers.
However, it did not converge in a reasonable time even when we prepared only one image patch per training sample.
Thus, we omitted its results.

\begin{figure*}[p]
	\centering
	\begin{tabular}{cc}
		\includegraphics{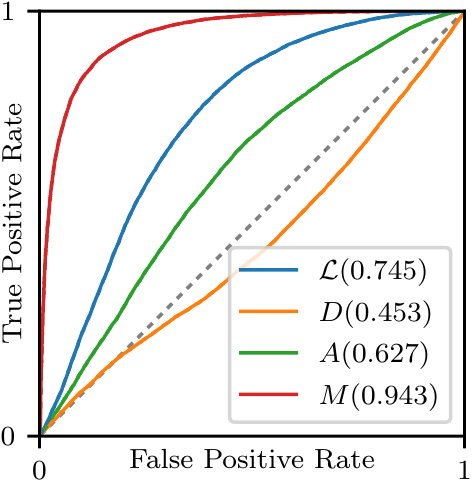}&
		\includegraphics{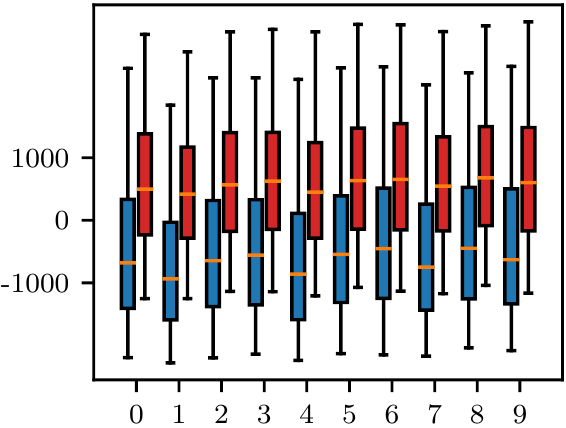}\\[-1mm]
		ROC curve&
		$\mathcal L$\\[2mm]
	\end{tabular}
	\begin{tabular}{ccc}
		\includegraphics{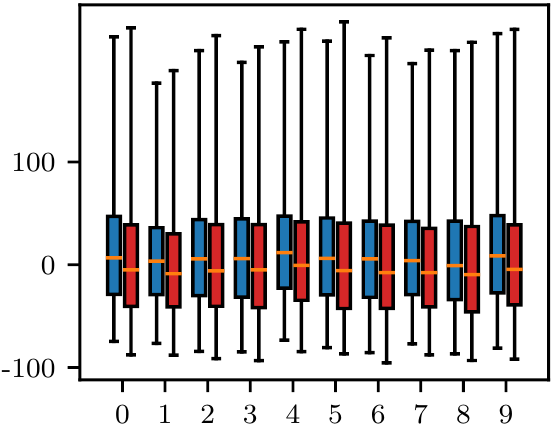}&
		\includegraphics{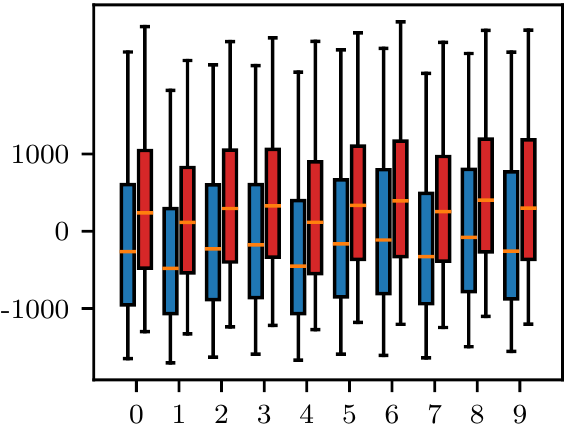}&
		\includegraphics{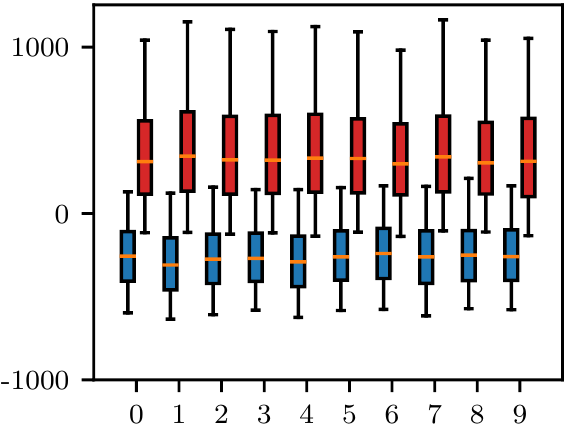}\\[-1mm]
		$D(x)$&
		$A(x)$&
		$M(x)$\\[2mm]
	\end{tabular}
	\vspace*{-5mm}
	\caption{
	Anomaly detection of the toy dataset by the VAE.
	(upper left panel) The ROC curves.
	The ROC-AUCs are in parentheses.
	(upper right panel) Box plots of the ordinary anomaly scores $\mathcal L(x)$ on each digit after subtracting the median value.
	The blue (and red) bars denote the first and third quartiles of the anomaly scores $\mathcal L(x)$ of normal (and anomalous) samples.
	The lower and upper whiskers denote the 5 and 95 percentiles, respectively.
	The anomaly scores $\mathcal L(x)$ of normal and anomalous samples overlap each other.
	(bottom panels) Box plots of the anomaly scores, $D(x)$, $A(x)$, and $M(x)$.
	The regularization term $D(x)$ is almost negligible.
	The log-normalizing constant $A(x)$ is slightly higher for anomalous samples, but it also varies depending on the digit: it is lower for ``1'' and ``7'' and higher for ``6'' and ``8''.
	The square normalized distance $M(x)$ is sensitive to the anomalous samples and insensitive to the digit.
	}
	\label{fig:svhnVAEstat}
\vspace*{2mm}
	\centering
	\begin{tabular}{cc}
		\includegraphics{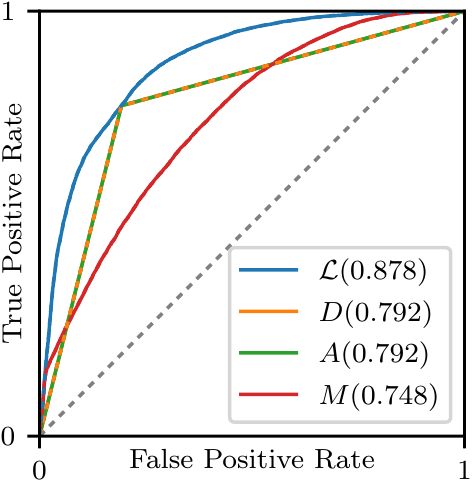}&
		\includegraphics{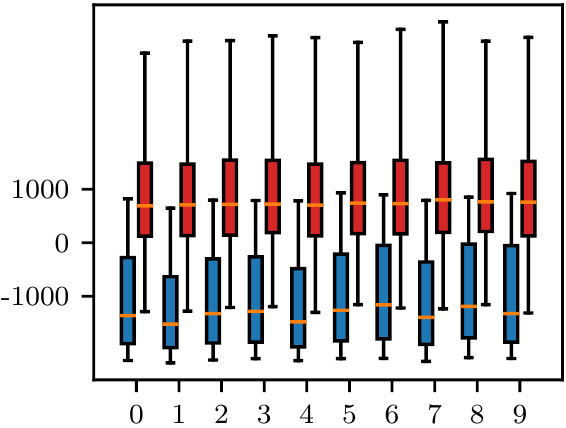}\\[-1mm]
		ROC curve&
		$\mathcal L$\\[2mm]
	\end{tabular}
	\begin{tabular}{ccc}
		\includegraphics{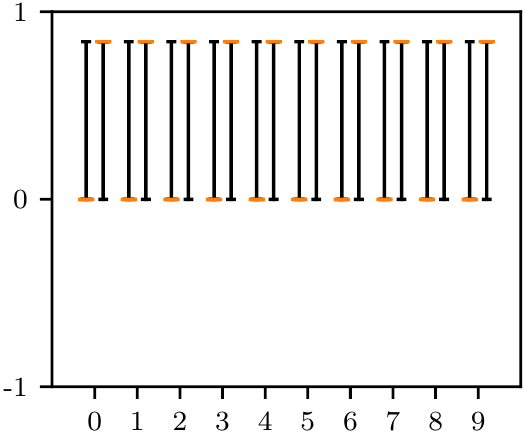}&
		\includegraphics{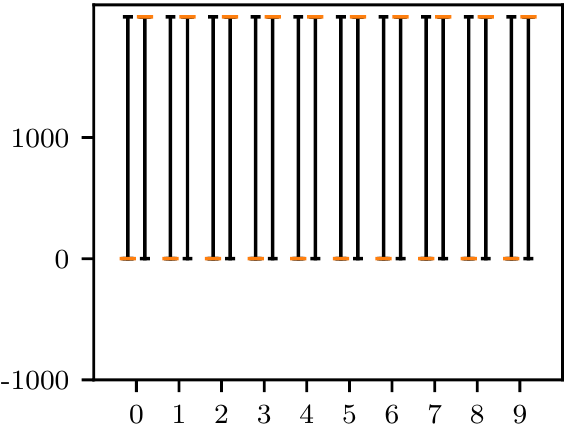}&
		\includegraphics{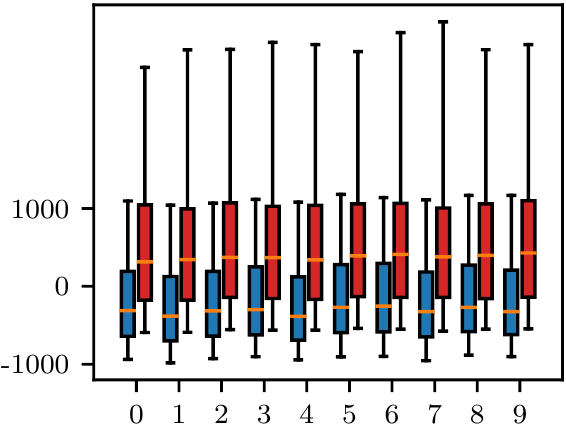}\\[-1mm]
		$D(x)$&
		$A(x)$&
		$M(x)$\\[2mm]
	\end{tabular}
	\vspace*{-5mm}
	\caption{
	Anomaly detection of the toy dataset by the GMM.
	The conditions are the same as those in Fig.~\ref{fig:svhnVAEstat} unless otherwise stated.
	(upper left panel) The ROC curves.
	The ROC-AUCs are in parentheses.
	(upper right panel) Box plots of ordinary anomaly scores $\mathcal L(x)$ on each digit after subtracting	the median value.
	(bottom panels) Box plots of the anomaly scores, $D(x)$, $A(x)$, and $M(x)$.
	Unlike the VAE, all three alternative anomaly scores, $D(x)$, $A(x)$, and $M(x)$, worked well and the ordinary anomaly score $\mathcal L(x)$ worked the best.
	}
	\label{fig:svhnGMMstat}
\end{figure*}

\begin{figure}[t]
	\centering
	\includegraphics[width=1.5in]{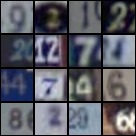}\hspace*{8mm}
	\includegraphics[width=1.5in]{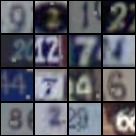}
	\caption{Reconstructions (mean values $\mu_x$) of the samples in Fig.~\ref{fig:svhn} by the VAE; (left panel) normal samples and (right panel) anomalous samples.
	}
	\label{fig:svhnVAErec}
\end{figure}

\begin{figure}[t]
	\centering
	\includegraphics[width=1.5in]{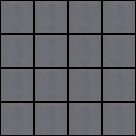}\hspace*{8mm}
	\includegraphics[width=1.5in]{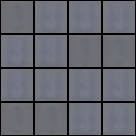}
	\caption{Reconstructions (mean values $\mu_k$) of the samples in Fig.~\ref{fig:svhn} by the GMM; (left panel) normal samples and (right panel) anomalous samples.
	}
	\label{fig:svhnGMMrec}
\end{figure}

\subsection{Results of Anomaly Detection}
Let $\mathrm{TP}$, $\mathrm{TN}$, $\mathrm{FP}$, and $\mathrm{FN}$ denote a true positive, true negative, false positive, and false negative, respectively.
We used the true positive rate ($\mathrm{TPR}$, or sensitivity) and the false positive rate ($\mathrm{FPR}$, or 1-specificity), defined as
\begin{align*}
  \mathrm{TPR}&=\mathrm{TP / (TP + FN)},\\
  \mathrm{FPR}&=\mathrm{FP / (FP + TN)}.
\end{align*}
We plotted the receiver operating characteristic (ROC) curves, which shows the relationship between $\mathrm{TPR}$ and $\mathrm{FPR}$ with a varying threshold.
We also calculated the areas under the ROC curves (ROC-AUCs).

\begin{figure}[t]
	\centering
	\includegraphics{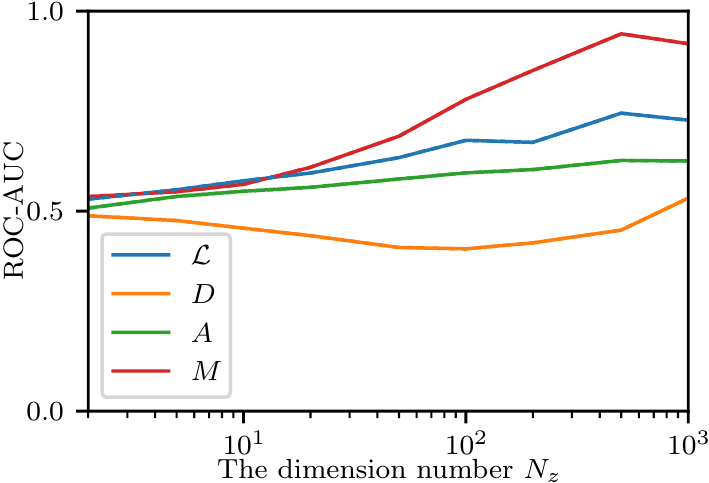}\\[1mm]
	\includegraphics{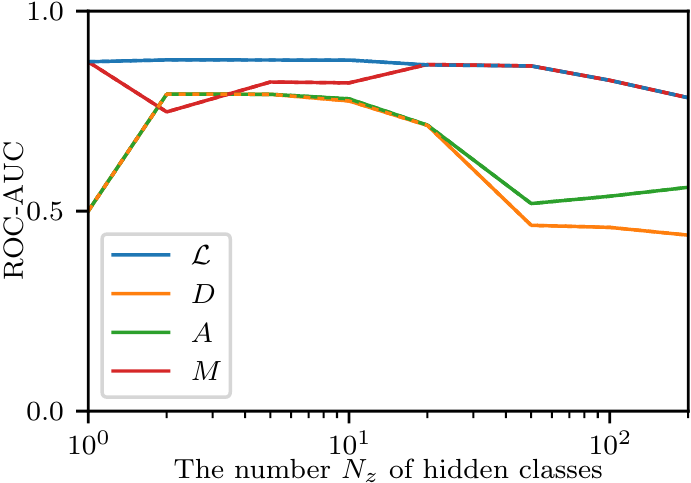}
	\caption{ROC-AUCs on the toy dataset with the varying hyperparameter.
	(top panel) VAE with the varying dimension number $N_z$ of the latent space $\mathcal Z$.
	(bottom panel) GMM with the varying number $N_z$ of hidden classes.
	}
	\label{fig:svhnROCAUCs}
	\vspace*{-2mm}
\end{figure}

For the toy dataset, we found that the VAE with $N_z=500$ achieved the best ROC-AUC of 0.745.
In that case, the VAE with the unregularized score $M(x)$ achieved an even better ROC-AUC of 0.943.
We summarize the ROC curves and the box plots of the anomaly scores in Fig.~\ref{fig:svhnVAEstat}.
The GMM achieved the best ROC-AUC of 0.878 with $N_z=2$ as summarized in Fig.~\ref{fig:svhnGMMstat}.
Figs.~\ref{fig:svhnVAErec} and~\ref{fig:svhnGMMrec} respectively depict the reconstructions (i.e., the estimated mean vector $\mu_x$) by the VAE and by the GMM of the original samples in Fig.~\ref{fig:svhn}.
Fig.~\ref{fig:svhnROCAUCs} depicts the resultant ROC-AUCs with the varying hyperparameter (the dimension number $N_z$ for the VAE, and the number $N_z$ of hidden classes for the GMM).

\begin{figure*}[t]
	\centering
	\includegraphics[width=3.0in]{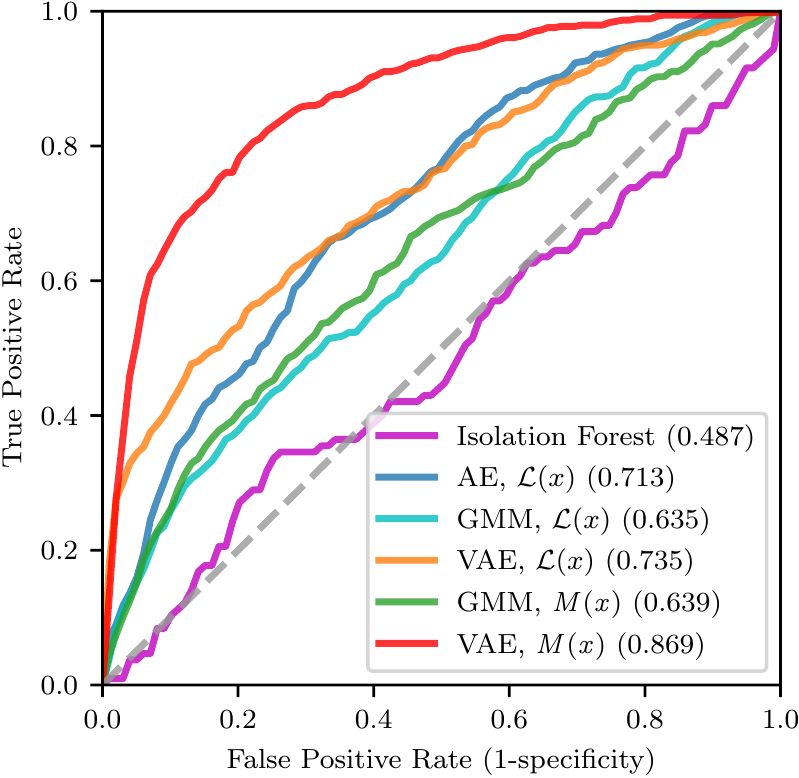}\hspace*{1cm}
	\includegraphics[width=3.0in]{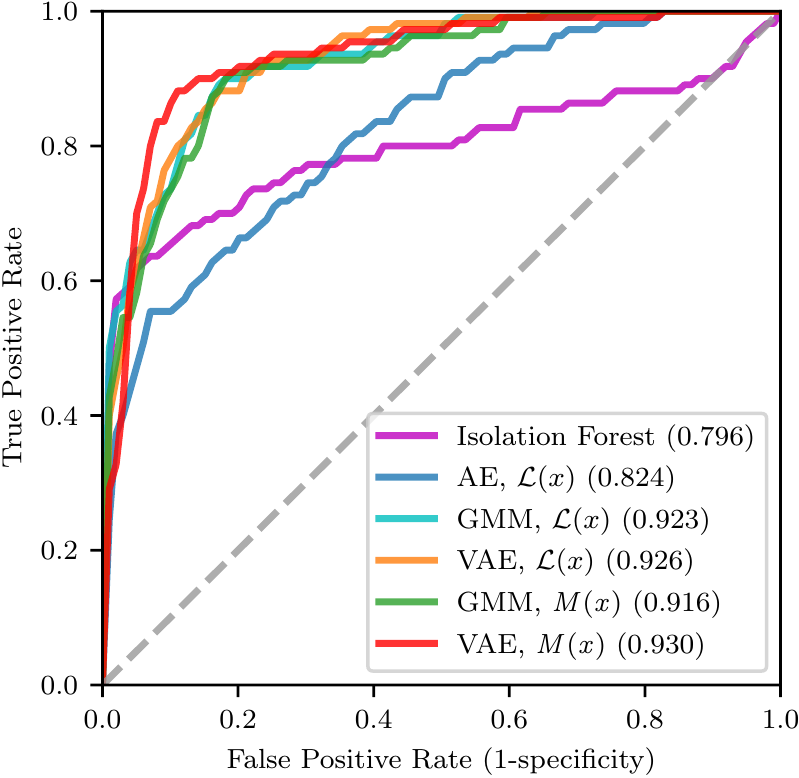}
	\vspace*{-3mm}
	\caption{
	Receiver operating characteristic (ROC) curves of the models with the best hyper-parameters.
	(left panel) Screw dataset.
	(right panel) Gear dataset.
	}
	\label{fig:roc}
\end{figure*}

We performed five trials for the screw dataset and a 5-fold cross validation for the gear dataset.
We show the best ROC-AUCs and the corresponding hyper-parameters in Table~\ref{tab:realaccuracy}.
Fig.~\ref{fig:roc} plots their ROC curves in the real-world manufacturing datasets.

\begin{table*}[t]
	\caption{Resultant ROC-AUCs in the Real-World Manufacturing Datasets.}
	\label{tab:realaccuracy}
	\centering
	\begin{tabular}{lc*{5}{c}}
	\toprule
									 &&\multicolumn{2}{c}{\textbf{Screw Dataset}} && \multicolumn{2}{c}{\textbf{Gear Dataset}} \\
	\cmidrule{3-4}\cmidrule{6-7}
	\textbf{Model} & \textbf{Score}        & \textbf{Hyper-Parameters} & \textbf{ROC-AUC} && \textbf{Hyper-Parameters} & \textbf{ROC-AUC} \\
	\midrule
		Isolation Forest~\cite{Liu2012}   & --- & $N_e=100,r=0.05$      & 0.487 && $ N_e=100,r=0.02 $      &  0.796 \\
		AE     & $\mathcal L(x)$     & $N_c=32,N_z=50$              & 0.713 && $ N_c=32, N_z=5 $ & 0.824 \\
	\midrule
	\multicolumn{2}{l}{\!\!\!Ordinary Anomaly Score}\\[.5mm]
		GMM    & $\mathcal L(x)$     & $N_h=200,N_z=10$             & 0.635 && $ N_h=100,N_z=20 $ & 0.923 \\
		VAE    & $\mathcal L(x)$     & $N_c=32,N_z=20$              & 0.735 && $ N_c=32,N_z=5 $ & 0.926 \\
	\midrule
	\multicolumn{2}{l}{\!\!\!Unregularized Score}\\[.5mm]
		GMM    & $M(x)$ & $N_h=200,N_z=10$    & 0.639 && $N_h=100,N_z=20$ & 0.916 \\
		VAE    & $M(x)$ & $N_c=32,N_z=50$    & \textbf{0.869} && $N_c=32,N_z=10$ & \textbf{0.930} \\
	\bottomrule
	\end{tabular}
\end{table*}

\section{Discussion}\label{sec:discussion}

\subsection{Anomaly Detection of the Toy Dataset}
As shown in the upper-left panel of Fig.~\ref{fig:svhnVAEstat} and the upper panel of Fig.~\ref{fig:svhnROCAUCs}, the VAE with the unregularized score $M(x)$ achieved a better ROC curve than the VAE with the ordinary anomaly score $\mathcal L(x)$.

While the ordinary anomaly score $\mathcal L(x)$ is slightly higher for anomalous samples, the score varies depending on the digit: it is lower for simply shaped digits like ``1'' and ``7'' and higher for complexly shaped digits like ``6'' and ``8'' (see the upper-right panel).
Thanks to its ability for rich expression, the VAE tried to reconstruct the details of each sample even when anomalous, as shown in Fig.~\ref{fig:svhnVAErec}.
The VAE imperfectly reconstructed the detailed features of normal samples (such as serifs and slight translations) as well as the anomalous regions (the gray squares).
Hence, the VAE with an ordinary anomaly score $\mathcal L(x)$ is sensitive both to anomalies and to the complexity of the samples.

The regularization term $D(x)$ negatively influences anomaly detection.
The encoder of the VAE ignores features of the anomaly, as described in Section~\ref{sec:concept}, and the anomaly obscures original features of a sample.
Hence, an anomalous sample provides a limited set of features, resulting in a non-characteristic latent variable (i.e., a latent variable close to the coordinate origin) and a smaller regularization term $D(x)$.
Leastwise, the regularization term $D(x)$ has the lowest score, one that is almost negligible.

The log-normalizing constant $A(x)$ has the largest variance among the three alternative scores and dominates the ordinary anomaly score $\mathcal L(x)$.
For regions that the VAE cannot reconstruct well, the decoder outputs the large log-normalizing constant $A(x)$ to prevent the square normalized distance $M(x)$ from exploding.
As a result, the log-normalizing constant $A(x)$ is sensitive to the complexity of samples as well as to anomalies.

Conversely, the square normalized distance $M(x)$ (i.e., the unregularized score) is sensitive to anomalous samples yet insensitive to the digit (i.e., the complexity of samples).
This is because it is normalized by the posterior variance $\sigma_x^2$.
We thus conclude that the unregularized score $M(x)$ is suited for anomaly detection in the case of heterogeneous complexity.

\subsection{Comparison with GMM on the Toy Dataset}
The GMM shows a different tendency in Fig.~\ref{fig:svhnGMMstat} and the bottom panel of Fig.~\ref{fig:svhnROCAUCs}.
All three alternative scores, $D(x)$, $A(x)$, and $M(x)$, worked well and the ordinary anomaly score $\mathcal L(x)$ worked the best.
The GMM outperformed the VAE with the ordinary anomaly score $\mathcal L(x)$ (see also Fig.~\ref{fig:svhnVAEstat} for comparison).

The GMM has a limited expression ability compared to the VAE, and it did not build a model of each sample, as shown in Fig.~\ref{fig:svhnGMMrec}.
However, the GMM built an anomaly class and successfully classified the anomalous samples.

Fig.~\ref{fig:svhnROCAUCs} shows that the ROC-AUCs worsened with the increase in the number $N_z$ of hidden classes.
In particular, the ROC-AUCs obtained using the negative log-mixture weight $D(x)$ and the log-normalizing constant $A(x)$ came close to the chance level of $0.5$.
This is because the GMM built several specific classes with lower mixture weights $w_z$ for sample clusters and produced more variations in log-normalizing constant $A(x)$, and the GMM misclassified samples in these classes as anomalous.
Since the VAE employs the continuous latent variable $z$, the VAE can be considered a mixture of an infinite number of hidden classes.
Given many hidden classes, the GMM might have encountered the same issue.
Thus, the proposed unregularized score might be useful for a GMM with much more hidden classes or for more diverse datasets.

Contrary to these results, previous studies reported remarkable performance with the VAE for anomaly detection~\cite{Zhai2016,Suh2016,Lopez-Martin2017,Chalapathy2017,Zhou2017,Ribeiro2017}.
They often evaluated models on a dataset of the novelty detection; for example, Chalapathy \emph{et al.}~\cite{Chalapathy2017} removed cat images from the training set of the CIFAR-10 dataset~\cite{cifar} and used cat images in the test set as anomalous samples.
In this case, the VAE does not learn anomalous samples, resulting in extremely high anomaly scores (i.e., reconstruction errors) and good detection performance.
The GMM has poor expression ability and reconstructs samples in known and novel classes at similar levels.
Then, the VAE outperforms the GMM and other competitive models.
On the other hand, we employed a training set that contained a limited number of anomalous samples, with an appearance that was similar to normal samples but with small scratches and cracks.
Consequently, and further confirmed by~\cite{Aytekin2018,Chen2018c} and in Fig.~\ref{fig:svhnVAErec}, the VAE partially reconstructed anomalous samples, resulting in only moderately high anomaly scores.
The increase in the anomaly scores of the anomalous samples is less significant than the variability in the anomaly scores among normal samples.
This is why the VAE did not perform as remarkably in our experimental setting and in other studies~\cite{Aytekin2018,Chen2018c}.


\begin{figure*}[t]
	\centering
	\begin{tabular}{cc}
	\includegraphics[width=1.72in]{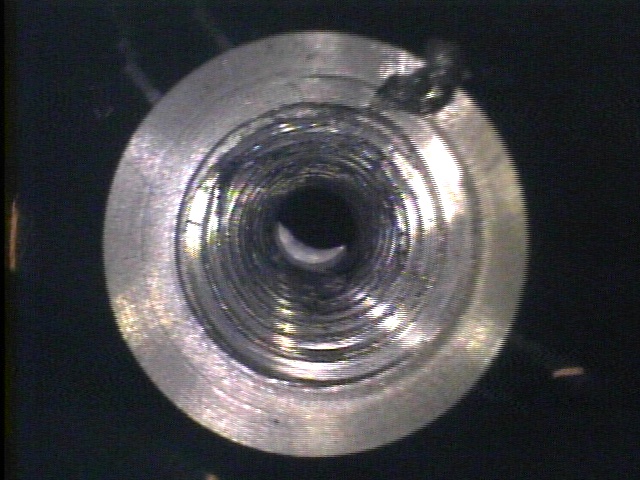}&
	\includegraphics[width=1.72in]{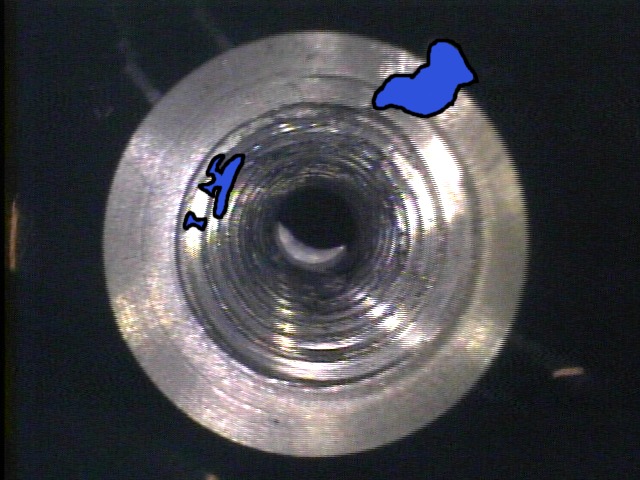}\\
	sample & ground truth\\[4mm]
	\end{tabular}
	\begin{tabular}{ccc}
		\includegraphics[width=1.72in]{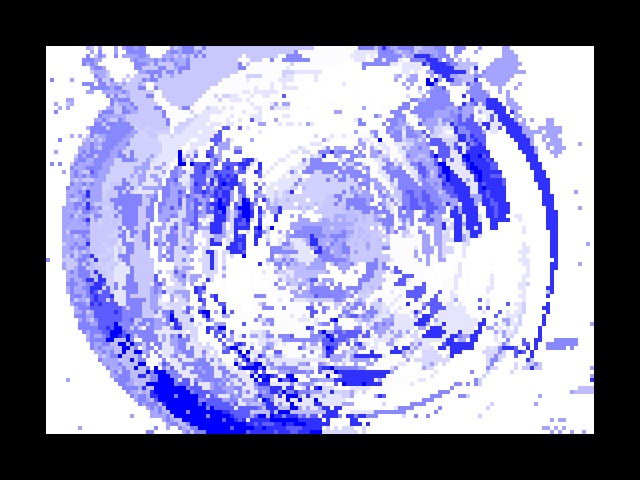}&
	\includegraphics[width=1.72in]{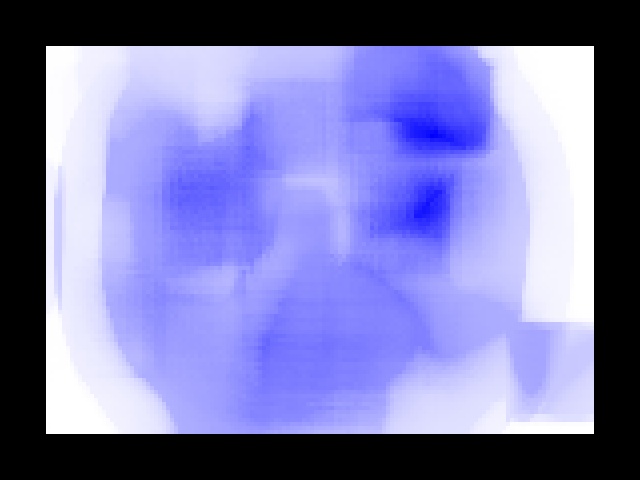}&
	\includegraphics[width=1.72in]{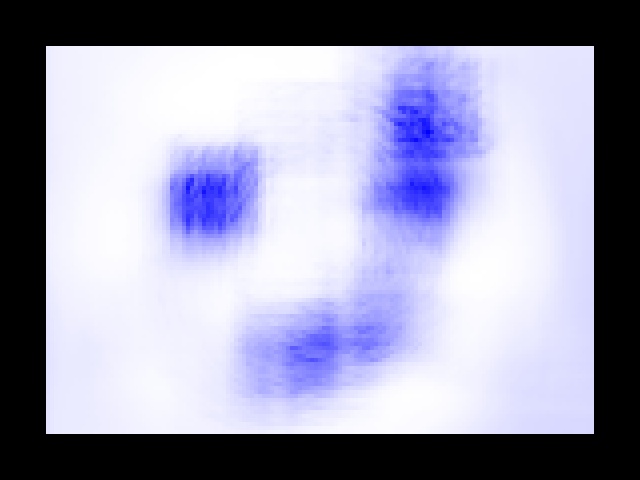}\\
	Isolation Forest $\mathcal L(x)$ &
	GMM $\mathcal L(x)$ &
	AE  $\mathcal L(x)$ \\[4mm]
	\end{tabular}
	\begin{tabular}{cccc}
	\includegraphics[width=1.72in]{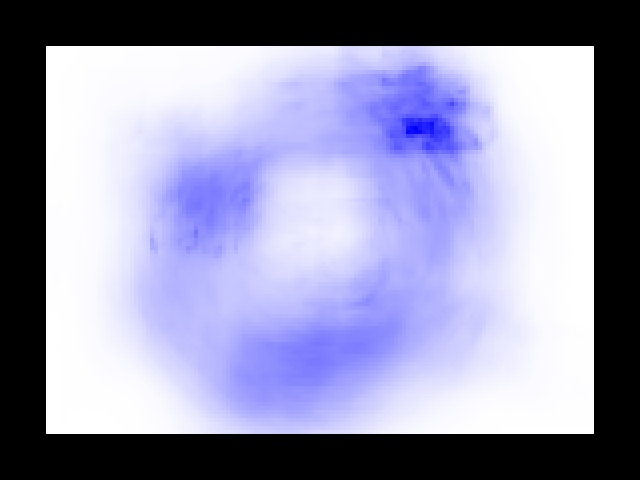}\hspace*{-2mm}&\hspace*{-2mm}
	\includegraphics[width=1.72in]{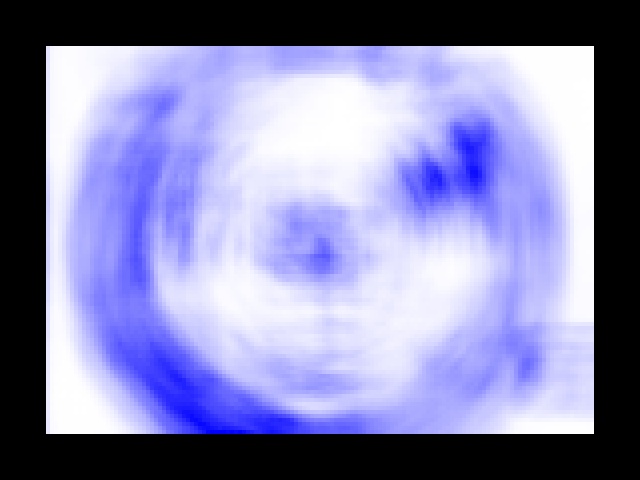}\hspace*{-2mm}&\hspace*{-2mm}
	\includegraphics[width=1.72in]{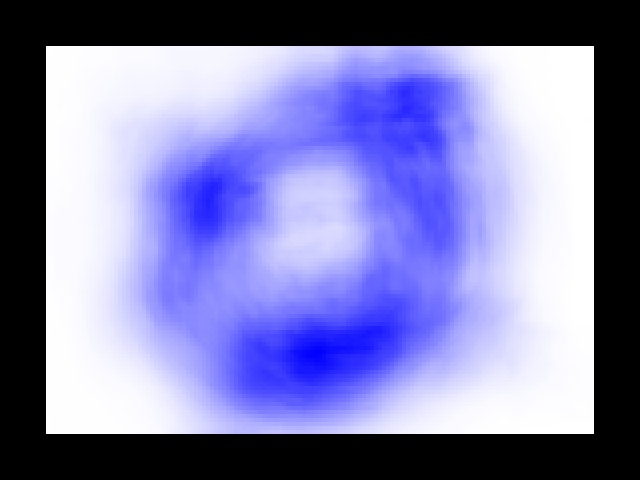}\hspace*{-2mm}&\hspace*{-2mm}
	\includegraphics[width=1.72in]{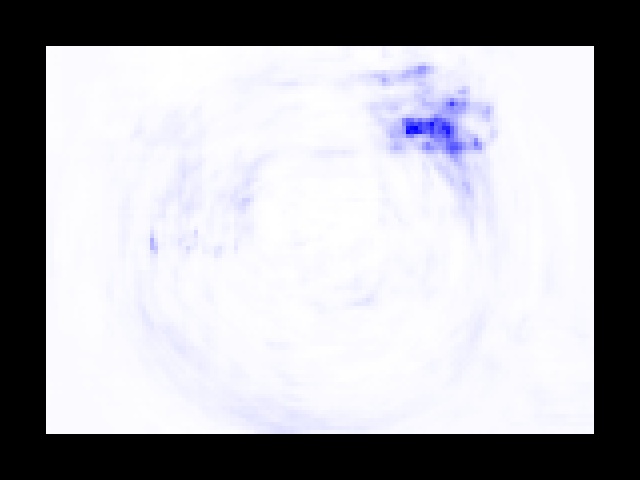}\\
	VAE $\mathcal L(x)$ &
	VAE $D(x)$ ($7.2 \times 10^1$)&
	VAE $A(x)$ ($2.3 \times 10^5$)&
	VAE $M(x)$ ($1.4 \times 10^5$)\\[4mm]
	\end{tabular}
	\caption{Example of an anomalous screw hole, the ground truth of anomalous areas, and the corresponding heat maps of the anomaly scores.
	The anomaly scores were normalized to the range of $[0,1]$.
	The scales of the alternative anomaly scores of the VAE are in parentheses.
	}
	\label{fig:heatmap}
\end{figure*}

\subsection{Anomaly Detection of Manufacturing Datasets}

As summarized in Table~\ref{tab:realaccuracy}, the VAE with the unregularized score $M(x)$ outperformed its counterpart by a large margin with the screw dataset.
Although it was not the aim of our study, we found that the GMM with the unregularized score $M(x)$ achieved a moderately better performance than its counterpart.
This might be because the screw dataset is more diverse than the toy dataset as described in the previous section.
For the gear dataset, despite the concern that the assumption in Section~\ref{sec:concept} might not hold, the proposed unregularized scores did not degrade the performance.

Fig.~\ref{fig:heatmap} shows a screw hole and the ground truth of anomalous areas.
One can find the cracks (known as ``blowholes'') in the upper-right part of the silver flat surface and in the left of the middle screw groove.
We also show the anomaly scores obtained by the comparative models with a stride of $4\times 4$, where each anomaly score is normalized to the range from 0 to 1 with brighter pixels representing a lower score.

As the ROC-AUC suggests, the Isolation Forest failed to build a model of the screw dataset.

The anomaly score $\mathcal L(x)$ of the GMM shows that the GMM is sensitive to the screw groove area and boundary areas between the silver and black surfaces as well as anomalous areas.
The GMM parcellates an image into a finite number of classes and is not robust to small parallel shifts, unlike the AE and VAE.
Hence, the GMM cannot build a model of the screw grooves and boundary areas.
As such, it did not work as well as it did with the toy dataset, where each sample was center-aligned.

The AE showed high anomaly scores $\mathcal L(x)$ in the anomalous areas but also in the screw groove area reflecting illumination light.
Since the AE has a constant standard deviation $\sigma_{x}$, it is sensitive to pixel intensities rather than semantic anomalies.

The VAE showed lower ordinary anomaly scores $\mathcal L(x)$ in the black flat surface area surrounding the screw hole and in the bottom area of the screw hole.
The VAE showed higher scores in the screw groove area despite the fact that these areas are coincident with the standards.
This result is unsurprising because the image patches from the screw grooves have more variation depending on the relative positions of the image patches and the starting points of the screw grooves.
The ordinary anomaly score $\mathcal L(x)$ is more sensitive to the imaging direction and to illumination compared to the case of the flat areas.
Hence, the likelihood of each image patch from the screw groove areas is naturally lower.
In other words, the ordinary anomaly score $\mathcal L(x)$ is related to the complexity of the target areas rather than to the anomaly, as predicted and discussed in Section~\ref{sec:concept}.
As with the toy dataset, the regularization term $D(x)$ shows the opposite tendency and is almost negligible compared to the other terms, $A(x)$ and $M(x)$.
The log-normalizing constant $A(x)$ has a dominant effect on the ordinary anomaly score $\mathcal L(x)$ and shows the same tendency.
The square normalized distance $M(x)$ is almost constant regardless of the areas where the image patches come from, provided that the areas are normal, and it produces high scores only for the anomalous areas.
The absolute error $|x-\mu_x|$ is high in the screw groove areas, but the squared normalized distance $M(x)$ is normalized by the standard deviation $\sigma_{x}$, which is related to the complexity of target areas, like the log-normalizing constant $A(x)$.
As a result, the square normalized distance $M(x)$ is robust to the complexity of the target areas.
For essentially the same reason, the square normalized distance $M(x)$ detected anomalous areas selectively.
Therefore, the proposed unregularized score $M(x)$ improves the accuracy of anomaly detection with machine components that have complex structures.

\section{Conclusion}
This study proposed an unregularized score using the variational autoencoder for anomaly detection.
As its name implies, the unregularized score is the objective function of a generative model without the regularization terms.
The variational autoencoder with the unregularized score was more robust to heterogeneously complex datasets, viz., datasets where the differences between simply and complexly shaped samples are higher than between normal and anomalous samples.
Moreover, the unregularized score worked well even for a simpler dataset.
Other datasets and other structured deep generative models will be explored in future work.

\section*{Acknowledgment}
The authors would like to thank the collaborators from AISIN AW CO., LTD.~for insightful suggestions regarding image data.
This study was partially supported by AISIN AW CO., LTD.~and the MIC/SCOPE \#172107101.



%


\end{document}